\documentclass[10pt,journal,compsoc]{IEEEtran}

\ifCLASSOPTIONcompsoc
  \usepackage[nocompress]{cite}
\else
  \usepackage{cite}
\fi
  \usepackage{amsmath,amssymb,amsfonts}
  \usepackage{algorithmic}
  \usepackage{graphicx}
  \usepackage{threeparttable}
  \usepackage{booktabs}
  \usepackage{float}
  \usepackage[linesnumbered,boxed,ruled,commentsnumbered]{algorithm2e}

\ifCLASSINFOpdf
\else
\fi
\begin{document}
%
\title{IUP: An Intelligent Utility Prediction Scheme for Solid-State Fermentation in 5G IoT}

\author{Min~Wang,
	Shanchen~Pang,
	Tong~Ding, Sibo~Qiao, Xue~Zhai, Shuo~Wang ,Neal N. Xiong~\IEEEmembership{Senior Member,~IEEE,} Zhengwen Huang  

	\IEEEcompsocitemizethanks{\IEEEcompsocthanksitem M. Wang is with Department of Control Science and Engineering, China
		University of Petroleum, Qingdao 266580, China.\protect E-mail: minwang2020@qq.com.
		
		\IEEEcompsocthanksitem S. Pang, S. Qiao and X. Zhai are with Department of Computer Science and Technology, China University of Petroleum, Qingdao 266580, China. \protect E-mail: pangsc@upc.edu.cn, 1175928825@qq.com.
		
		\IEEEcompsocthanksitem T. Ding is with school of software, Shandong University, China. \protect E-mail: 793394086@qq.com.
		
		\IEEEcompsocthanksitem Neal N. Xiong is with Department of Mathematics and Computer Science
		Northeastern State University, Tahlequah, OK, USA. \protect\\
		E-mail: xiongnaixue@gmail.com, xiong31@nsuok.edu.

		\IEEEcompsocthanksitem S. Wang is Department of Computer Science and Technology, Tongji University. \protect E-mail: 8362084547@qq.com. 
		
	}
}

\markboth{Journal of \LaTeX\ Class Files,~Vol.~14, No.~8, August~2015}%
{Shell \MakeLowercase{\textit{et al.}}: Bare Advanced Demo of IEEEtran.cls for IEEE Computer Society Journals}
%



\IEEEtitleabstractindextext{%
\begin{abstract}
At present, SOILD-STATE Fermentation (SSF) is mainly controlled by artificial experience, and the product quality and yield are not stable. Accurately predicting the quality and yield of SSF is of great significance for improving human food security and supply. In this paper, we propose an Intelligent Utility Prediction (IUP) scheme for SSF in 5G Industrial Internet of Things (IoT), including parameter collection and utility prediction of SSF process. This IUP scheme is based on the environmental perception and intelligent learning algorithms of the 5G Industrial IoT. We build a workflow model based on rewritable petri net to verify the correctness of the system model function and process. In addition, we design a utility prediction model for SSF based on the Generative Adversarial Networks (GAN) and Fully Connected Neural Network (FCNN). We design a GAN with constraint of mean square error (MSE-GAN) to solve the problem of few-shot learning of SSF, and then combine with the FCNN to realize the utility prediction (usually use the alcohol) of SSF. Based on the production of liquor in laboratory, the experiments show that the proposed method is more accurate than the other prediction methods in the utility prediction of SSF, and provide the basis for the numerical analysis of the proportion of preconfigured raw materials and the appropriate setting of cellar temperature.
\end{abstract}

}

\maketitle

\IEEEdisplaynontitleabstractindextext

%
\IEEEpeerreviewmaketitle

\ifCLASSOPTIONcompsoc
\IEEEraisesectionheading{\section{Introduction}\label{sec:introduction}}
\else
\section{Introduction}
\label{sec:introduction}
\fi

%
%
%
%
\IEEEPARstart{S}{olid-state} fermentation (SSF) has been defined as the fermentation process which involves solid matrix and is carried out in absence or near absence of free water [1]. The purpose of SSF is to accumulate the target metabolites. SSF takes a certain proportion of raw materials and an appropriate cellar-entry temperature as the main preconditions for cellar-entry fermentation [2]. Once the raw material was sent into the fermentation cellar, no operation can be applied to the fermentation cellar until the end of fermentation. The yield and quality of SSF always instability by traditional method which depends on artificial expertise to control the proportion of raw materials. The relationship between raw material parameters should be analyzed. The 5G Internet of Things (IoT) is one of the technologies to realize industrial mass production, which provide the method to collect the parameters [3]. Therefore, the traditional industry of SSF should be further enhanced with the technology of 5G IoT to realize the intelligent manufacturing of SSF industry. The production process of Chinese liquor is a typical SSF [4]. We study the utility prediction of the fermentation process of Chinese liquor, which provide a method to predict the yield and quality of SSF.

Now, changing raw materials and adjusting cellar temperature are the two main method to improve the Chinese liquor product quality and yield of SSF. [5] analyzes the influence of different raw materials on the quality but the effects of different proportions of materials on liquor quality is not studied. Microbial is another factors which affect the quality of SSF. [6][7] research the quality of liquor of different microbial community. The main microorganisms affecting the quality of rice-liquor were analyzed in [8], and the optimum temperature of the microorganism was studied. Changes in the temperature of the grains in the fermentation cellar will have a significant impact on the growth and metabolism of microorganisms, which in turn will affect the yield and quality of liquor [9][10]. In [11], authors research the influence of the temperature trend in the fermentation cellar on the yield and quality of liquor, and it is concluded that the temperature curve of high-quality liquor should be with the trend of rise slowly in the early stage, rise rapidly in the middle stage and drop slowly in the later stage.

Nevertheless, different liquors have different microbial communities. The microorganism through biological experiments, especial the impact of microbes on the liquor SSF is a essential research field. At the same time, the research of real-time monitoring and control of the fermentation temperature ignores that fermentation cellar cannot be opened during the SSF process. In order to improve the yield and quality of all liquors without controlling the fermentation process, another effective method is to control the key preconditions of SSF in advance, because the proportion of preconditions is a vital factor affecting the quality and yield of SSF products [11]. 

It is a technical approach that improve the utility of SSF by optimizing the key cellar entry preconditions. IIoT provides the necessary methods for collecting parameters such as liquor Alcohol, Temperature, humidity, Starch content, and Acidity in the SSF process. These parameters provide the data analysis samples for the optimization of key cellar-entry preconditions of SSF. Parameters collection and prediction system of SSF of Chinese liquor based on 5G Industrial IoT is shown in Fig. 1. The system analytical model based on the 5G Industrial IoT established in this paper plays a key role in analyzing the mathematical relationship between parameters and predicting the quality and yield of products correctly. Moreover, it is necessary to verify correctness of the model.

Since the parameters of SSF process are obtained by regular or irregular collection, which is a typical dynamic discrete event, and the collection times of parameter samples are dynamic. Hence, the system model needs reconstruction in the collection structure. Petri net is a graphic modeling tool with a strict mathematical definition and is well applied to describe the process such as discrete, synchronous, asynchronous, and concurrent processes [12]. Rewritable Petri nets [13] was proposed to solve the formal description and modeling in dynamic system reconstruction. Rewritable Petri nets provide a better analysis and verification method for dynamic discrete systems [14][15] with structural reconstruction. Therefore, the rewritable petri net can be well applied to model and verify the system model of the parameter collection and analysis of SSF. To predict the quality and yield of SSF, we use deep learning to analysis the hidden relationship between parameters. Neural network training needs numerous data. More sufficient data make the relationship mining more accurate, which leads to more ways to generate more data.

In this work, Our main contributions are as follows:

•	To establish an IUP framework of SSF of Chinese liquor based on deep learning. 

•	To guarantee  the correctness of the framework, we propose an edge-rewritable petri net to model and verify the soundness of the system. 

•	To realize the learning parameter relationship in the system model, we design an MSE-GAN model to expand the parameters of SSF process and analyze the relationship between these parameters of SSF process by using fully connected neural network. 

The rest of this work is arranged as follows: the related works are introduced in Section II. Section III introduces the system framework and builds the rewritable petri net model of the system framework, and analyzes the soundness of the model. Section IV describes the algorithm of MSE-GAN and the utility prediction model of SSF based on fully connected neural network. Section V verifies our method and experiments.  The last Section summarizes the work of the paper and future work.
\begin{figure}
	\centering
	\includegraphics[width=0.5\textwidth]{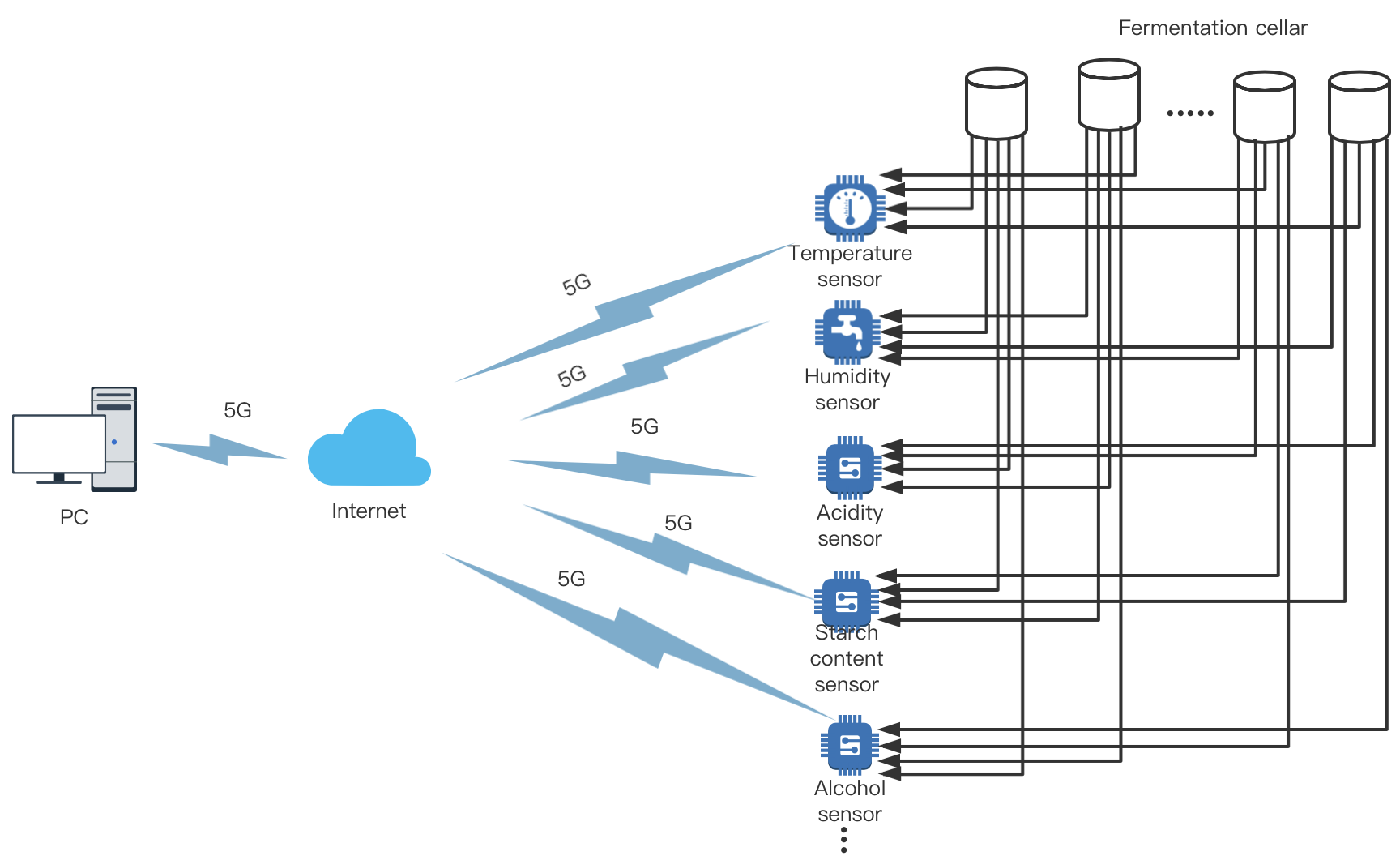}
	\caption{ The collection and prediction system of SSF process of Chinese liquor based on Internet of things.}
	\label{fig:example}
\end{figure}

\section{Related work}
The main raw material of SSF is grain. In 2019, the consumption of grain in the wine industry by SSF reached 30 million tons. The sauce, vinegar and other products in our daily life are all products of SSF. According to statistics, about 10\%-15\% of grain is converted into food necessities through SSF. This year, the impact of COVID-19 given high priority to food security, leading to the worst global food crisis in 50 years. Therefore, it is of great significance to predict the fermentation environment, establish the best SSF environment, and improve the yield and quality of SSF to ensure the national food security. At present, many researches focus on improving the quality and yield of SSF. In [42], the use of oilseed as the substrate for SSF can improve the content of unsaturated fatty acids and protein in the products. In [43], microorganisms are used to improve the production process of SSF, which can increase the yield of SSF and shorten the time of SSF process. Authors in [44] Response Surface Methodology (RSM) to optimize the parameters of SSF such as time, pH and temperature can enhance the production of alginate

The SSF is mainly optimized by microorganisms. However, each product of SSF needs to be analyzed because different raw materials use different microorganisms. As a result, microbial optimization of SSF is not universal. The intelligent production of SSF industry is realized by using IoT. The industrial IoT mainly uses sensor [45][46] technology to collect data in the industrial production process. In order to solve the security problem of industrial control system, Zhang et al [6] design a dynamic network security risk assessment method, which can assess the risk caused by unknown attacks. Authors in [47] develop a new security design using MODBUS protocol to provide higher security guarantee for IT infrastructure. Yang et al [48] design a robot obstacle avoidance algorithm, which is effectively applied in industrial automation production. In order to realize the efficient production in SSF industry, it is necessary to collect the data of the SSF process through the sensor in real time to analyze the relationship between the data by using the deep learning method. This will improve the yield and quality of the SSF.

The process of IUP of SSF is a workflow. It is necessary to model the workflow to ensure the correctness of it. The traditional workflow models include the  Event-Condition-Action (ECA), Business Process Execution Language (BPEL) and Yet Another Workflow Language (YAWL). These modeling approaches describe business process as a set of activities executed by a fixed control flow. In [31], An object-centered modeling method is proposed, which defines three abstract types of business objects responsible for creating and managing tasks and subprocesses, and modeling the life cycle of business objects by using gateway Extended Finite State Machine (FSM). However, the model constructed by the modeling method need to be conversed to YAWL model and performed in YAWL engine [32]. Lin et al. [33] model the serverless applications, predict the average end-to-end response time and workflow cost through the model to realize the optimal configuration of serverless applications. However, these model do not scale well and have some limitations when modeling dynamic systems. Petri net is a model developed to describe distributed system. In [30], authors use the petri net to model the application over the heterogeneous clouds to manage these applications. In [13], the authors propose a place rewritable petri net which can be used to model the reconfiguration systems. According to the process of collecting the parameters of SSF process, we propose the edge-rewritable petri net to model the SSF process system.

The data generated from the process of SSF is in one-dimensional format. We need to generate more effect data when the collected data is not sufficient. Generative Adversarial Networks (GAN) was proposed by Goodfellow in [18], which can generate effective data by using few-shot data. GAN is not relay on any prior assumptions in data generation. It makes the generated data conform to the distribution of actual data through the game between Generator (G) and Discriminator (D). The G network receives a random noise z, and generates data of approximate samples as much as possible. The D network receives an input data and tries to discriminate whether the data is a real sample or a fake sample generated by the network.           

Now, GAN is widely applied in the image data generation [19], voice data generation [20], text data generation [21], etc. The image data augmentation method of pedestrians of small scale or in heavy occlusions is proposed in [25], the image generated in this paper holds good visual quality as well as attributes. In [26], authors repair the aging image, the original image data is supplemented by GAN, and the high-quality image is generated. The image data is mainly stored in the form of matrix. For the generation of one-dimensional data, the storage mode of data should be converted. In [34], a Classification enhancement GAN is proposed to solve the problem of data imbalance in classification, which enhance the accuracy of target prediction in the case of data imbalance. Liu et al. [35] propose an Adversarial Symmetric GANs (AS-GANs) which combines the training to real samples by D network with ordinary GAN. It makes the adversarial training become symmetry, stable training and accelerate the convergence. Yin et al. [36] introduce the time series into GAN, which can effectively solve the problem of disharmony between intelligent generation control and generation command scheduling. In view of the successful application of GAN in image data processing, we apply GAN to the production of one-dimensional data. 

After obtaining more valid data, the relationship between different attribute data needs to be analyzed. Bayesian network (BN) models and the Markov model (MM) are two even prediction model. However, the prediction of these two methods mainly depends on the latest state. These methods can be used for reasoning and calculation, but the whole process is oversimplified and the important characteristics of many state nodes are ignored, resulting in low accuracy of prediction. FCNN is a computational model that imitates biological neural network [22], it has three parts: input layer, hidden layer and output layer. FCNN is a modeling technology of deep learning and is widely used in the field of classification, prediction etc. In [27], authors calculate the tilt misalignments of the off-axis telescope by FCNN. In [28], this paper classify radar clutter and real radar target by using the FCNN. [29] solves the accurate classification of the arrhythmia by a dual FCNN.

FCNN has a number of research achievements in the field of prediction. In the [23], authors use the LSTM to simulate the local variation of PM2.5 pollution and then use FCNN to predict the air quality. The FCNN is used to predict and identify the arc faults in [24]. In [29], authors use the FCNN to predict the environmental factors of landslide susceptibility. A global-to-local localization approach using FCNN is proposed in [37] to localize anatomical landmarks in medical images. In [38], The combination of FCNN and two LSTM components can effectively predict train delays caused by weather and other conditions. The FCNN is combined with VGG19 in [39], which covers the multiple scale features. As a result, it can effectively predict the degree of cancer lesions. Wang et al., [40] found that FCNN could measure the conditional correlation between genes by using the similarity of co-expression and prior knowledge, and the experiment proved that FCNN could effectively predict the genetic relationship.  Ale et al. [41] uses FCNN to learn and predict BRNN, which realizes the prediction of active cache of edge computing to relieve the pressure of core network in the Internet of Things. Based on these researches about the FCNN, FCNN can be well applied in analyzing and predicting data, so we adopt the FCNN to predict the quality and yield of SSF.

\section{System Model of Rewritable Petri Net}
In order to guarantee the correctness of the system framework of IUP of SSF. This Section employs the edge-rewritable petri net to model and illustrate the correctness of the system framework.

\subsection{System Framework}
The system framework include parameter collection of SSF process and study the relationship between parameters. The main parameters in SSF of Chinese liquor include the acidity, starch content, humidity, maternal draff, Daqu, bran shell, original cellar temperature and ground temperature. These parameters affect the cellar temperature and the quality and yield of liquor. The factors of acidity, starch content, humidity, maternal draff, Daqu, bran shell affect each other in pairs. In order to obtain the mathematical relationship between these parameters and realize the intellectualization of liquor industry, this paper combines the SSF industry with deep learning to realize the utility prediction of SSF. The system framework diagram is shown in Fig. 2. 

Step1: parameters will be collected several times by the sensor.

Step2: the relationship between parameters will be studied by deep learning.

Step3: the utility prediction model of SSF are obtained.

The utility prediction model provides a basis for the numerical analysis of the preconditions of SSF and provides a guidance for the regulation of microbial community, so as to optimize the liquor production. 
\begin{figure}
	\centering
	\includegraphics[width=0.5\textwidth]{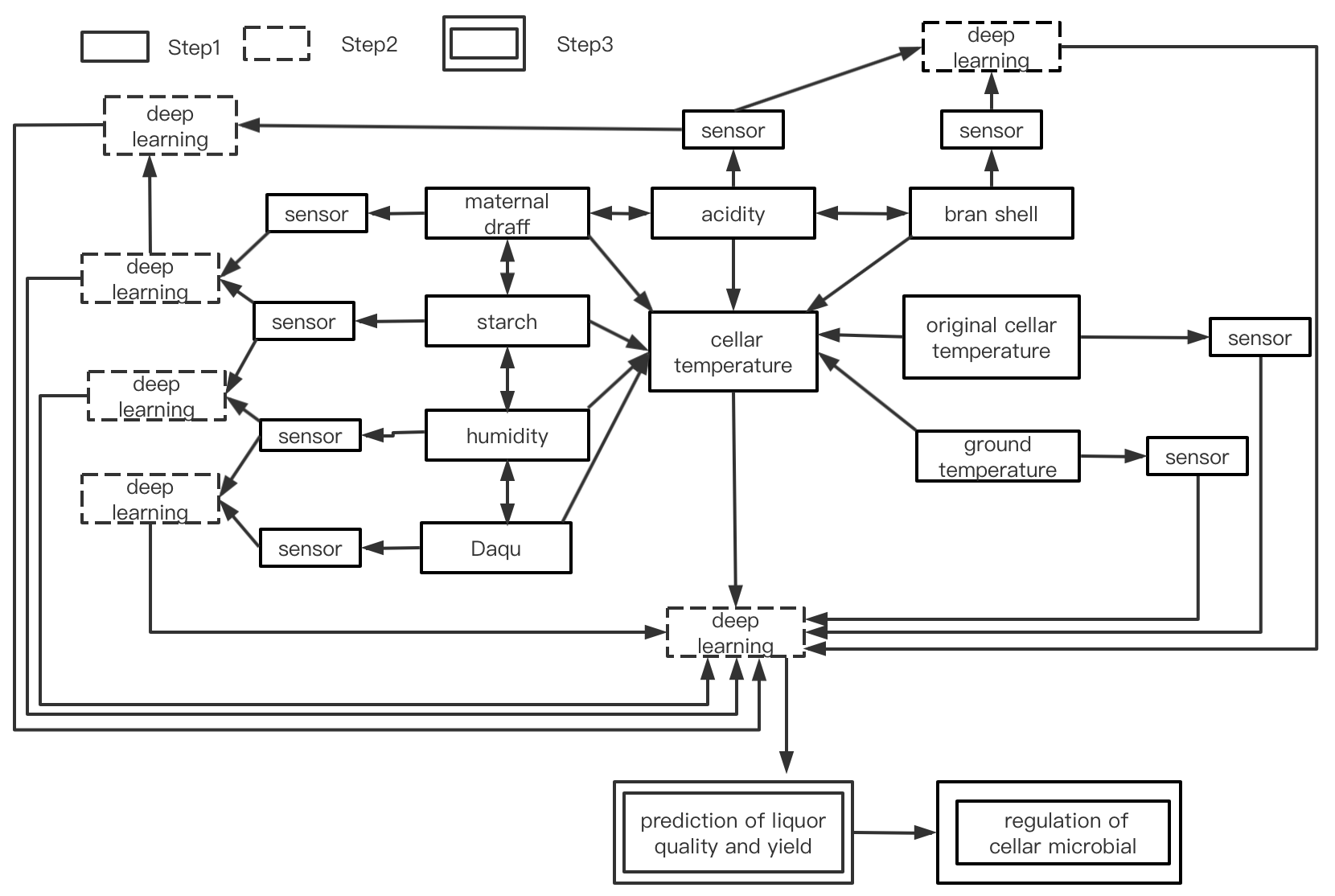}
	\caption{The system framework of IUP of SSF.}
	\label{fig:example}
\end{figure}

\subsection{Edge Rewritable Petri Net}
Petri net is a workflow modeling and analysis tool. According to the characteristic that the parameters of SSF process need to be collected several times, the rewritable Petri net is proposed to model the framework.

\textbf{Definition 1} (petri net[13]) The 4-tuple \emph{N} = (\emph{P}, \emph{T}, \emph{F}, \emph{M}$_{0}$) is called a Petri net , where
\emph{P} is a set of places.
Each place have several tokens, whose length of tuples is determinate.
\emph{T} is a finite set of transition.
\emph{P} $\cap$ \emph{T} = $\varnothing$.
Iff 3-tuple (\emph{P}, \emph{T}, \emph{F}) is a net.
\emph{F} $\subseteq$ (\emph{P} $\times$ \emph{T}) $\cup$ (\emph{T} $\times$ \emph{P}) is a set of arcs.
A  marking of \emph{N} is a function \emph{M}: \emph{P} $\rightarrow$ $\mathbb{N}$.
\emph{M}$_{0}$ is an initial marking, $\mathbb{N}$ = \{0, 1, 2,...\} is a set of non-negative integer.

(1) $^{.}$\emph{x} = \{\emph{y} $\mid$ (\emph{y}, \emph{x}) $\in$ \emph{F}\} is a input set or preset of \emph{x}.
And \emph{x}$^{.}$ = \{\emph{y} $\mid$ (\emph{y}, \emph{x}) $\in$ \emph{F}\} is called output set or post-set of \emph{x}.

(2) $\forall$ \emph{M}$_{1}$ and \emph{M}$_{2}$, \emph{M}$_{1}$ $\leq$ \emph{M}$_{2}$,
if and only if for all \emph{p} $\in$ \emph{P}: \emph{M}$_{1}$(\emph{p}) $\leq$ \emph{M}$_{2}$(\emph{p}).

Then the firing rule of N is introduced as follows:

(3) Transition \emph{t} $\in$ \emph{T} is enabled in a marking \emph{M}, iff $\forall$ \emph{p} $\in$ \emph{$^{.}$t},
\emph{M}(\emph{p}) $\geq$ 1, which is denoted as \emph{M}[\emph{t}$\rangle$.

(4) If transition \emph{t} $\in$ \emph{T} is enabled in a marking \emph{M}, the \emph{t} can be fired and generate a new marking \emph{M}$^{\prime}$,
which is denoted as \emph{M}[\emph{t}$\rangle$\emph{M}$^{\prime}$:
$$ \emph{M}'=\left\{
\begin{array}{rcl}
\emph{M}(\emph{p}) - 1,\hfill      &      & {\emph{p} \in \emph{$^{.}$t} - \emph{t$^{.}$}; }\\
\emph{M}(\emph{p}) + 1,\hfill     &      & {\emph{p} \in \emph{t$^{.}$} - \emph{$^{.}$t};}\\
\emph{M}(\emph{p}),\hfill         &      & {other.}\\
\end{array} \right. $$

Suppose that \emph{R}(\emph{M}$_{0}$) represents all marking set, which N\emph{} can reach from \emph{M}$_{0}$.
If $\exists$ \emph{M}$_{1}$, \emph{M}$_{2}$, ..., \emph{M}$_{k}$ $\in$ \emph{R}(\emph{M}$_{0}$) and $\forall$ \emph{i}:
1$\leq$\emph{i}$\leq$\emph{K}, $\exists$ \emph{t$_{i}$} $\in$ \emph{T}: \emph{M}$_{i}$[\emph{t}$_{i}$$\rangle$\emph{M}$_{i+1}$,
the transition sequence $\sigma$ = \emph{t}$_{1}$\emph{t}$_{2}$...\emph{t}$_{k}$ is enabled in a marking \emph{M}$_{1}$,
\emph{M}$_{k+1}$ is reachable from \emph{M}$_{1}$, which is denoted as \emph{M}$_{1}$[$\sigma$$\rangle$\emph{M}$_{k+1}$.

\textbf{Definition 2} (edge rewritable petri nets) A 7-tuple \emph{EN}=(\emph{P}, \emph{T}, \emph{F}, \emph{K},, \emph{W}, \emph{M}, \emph{W}$_{v}$) is an edge rewritable petri net, where (\emph{P}, \emph{T}, \emph{F}) is a basic petri net:

(1) \emph{W}:F$\rightarrow$\{0, 1, 2,...\} is a weight function, \emph{K}:S$\rightarrow$\{0, 1, 2,...\} is a capacity function, \emph{M}:P$\rightarrow$\{0, 1, 2,...\} is a marking function of \emph{EN} and satisfy the condition of $\forall$\emph{p} $\in$ \emph{P}: \emph{M}(\emph{s}) $\leq$ \emph{K}(\emph{s}).

(2) The firing rules of transition t in \emph{EN} is introduced as follows:
$$ \left\{
\begin{array}{rcl}
\emph{M}(\emph{p}) \geq \emph{W}(\emph{p,t}),\hfill      &      & {\emph{p} \in \emph{$^{.}$t} - \emph{t$^{.}$}; }\\
\emph{M}(\emph{p}) + \emph{W}(\emph{p,t})\geq \emph{K}(\emph{p}),\hfill      &      & {\emph{p} \in \emph{t$^{.}$} - \emph{$^{.}$t}; }\\
\emph{M}(\emph{p}) - \emph{W}(\emph{p,t})+\emph{W}(\emph{t,p})\geq \emph{K}(\emph{p}),\hfill      &      & {\emph{p} \in \emph{$^{.}$t} \cap \emph{t$^{.}$}. }\\
\end{array} \right. $$
If transition \emph{t} $\in$ \emph{T} is enabled in a marking \emph{M}, the \emph{t} can be fired and generate a new marking \emph{M}$^{\prime}$,
which is denoted as \emph{M}[\emph{t}$\rangle$\emph{M}$^{\prime}$:
$$ \emph{M}'=\left\{
\begin{array}{rcl}
\emph{M}(\emph{p}) - \emph{W}(\emph{p,t}),\hfill      &      & {\emph{p} \in \emph{$^{.}$t} - \emph{t$^{.}$}; }\\
\emph{M}(\emph{p}) + \emph{W}(\emph{p,t}),\hfill      &      & {\emph{p} \in \emph{t$^{.}$} - \emph{$^{.}$t}; }\\
\emph{M}(\emph{p}) - \emph{W}(\emph{p,t})+\emph{W}(\emph{t,p}),\hfill      &      & {\emph{p} \in \emph{$^{.}$t} \cap \emph{t$^{.}$}; }\\
\emph{M}(\emph{p}),\hfill         &      & {other.}\\
\end{array} \right. $$

(3) $\exists$\emph{f} $\in$ \emph{F} is an rewritable edge, when \emph{t}$_{i}$ $\in$ \emph{T} and \emph{t}$_{i}$ is a vertex of \emph{f}, \#(\emph{t}$_{i}$/$\sigma$)= \emph{W}$_{v}$$\rightarrow$ \emph{F} = \emph{F}-{\emph\{f\}}.  \emph{W}$_{v}$ $\in$ $\mathbb{N}$$^{*}$ is the rewritable restriction of  \emph{f}. $\sigma$ is a sequence of transitions,  \#(\emph{t}$_{i}$/$\sigma$) denotes the number of \emph{t}$_{i}$ occurrence in $\sigma$. The rewritable edges are represented by dashed lines.

\textbf{Definition 3}  \emph{N} = (\emph{P}, \emph{T}, \emph{F}) is a workflow[23] where

(1)	If \emph{N} has two special places start and end that \emph{$^{.}$\emph{start}}=$\varnothing$, \emph{\emph{start}$^{.}$}=$\varnothing$.

(2)	\emph{N} is an extended workflow with a new transition t that \emph{$^{.}$t}=\{\emph{start}\} and  \emph{t$^{.}$}=\{\emph{end}\} . This extended workflow is strongly connected.

\subsection{The rewritable petri net model of the system framework}
To analysis the relationship between the raw materials parameters and alcohol. We use the edge-rewritable petri net to model the framework. The place P denotes parameters and relation models of parameters, meanwhile, the transition T denotes the operation of collecting and studying the parameters relationships. The rewritable edges can realize the multiple collecting parameters.

According to the system framework, we use the rewritable petri net to model the system framework for IUP of SSF. Capital letters are used to indicate the parameters of Chinese liquor SSF. \textbf{M}: master grains, \textbf{S}: starch content, \textbf{D}: Daqu, \textbf{H}: humidity, \textbf{A}: acidity, \textbf{B}: bran shell, \textbf{C}: cellar temperature, \textbf{G}: ground temperature, \textbf{ALC}: alcohol. The rewritable petri net model of the system framework is shown in Fig. 3. The meaning of all the elements in Fig. 3 are shown in Table I and Table II.  

In this model, \emph{P}$_{start}$ is a start place with a token, transition \emph{t}$_{0}$ denotes the operation of taking materials into the cellar. First, \emph{t}$_{0}$ can be fired and these parameters places \emph{P}$_{B}$,\emph{P}$_{A}$,\emph{P}$_{M}$,\emph{P}$_{S}$,\emph{P}$_{H}$,\emph{P}$_{D}$,\emph{P}$_{C}$,\emph{P}$_{G}$,\emph{P}$_{ALC}$ obtain a token. Now, the collected transition of \emph{t}$_{1}$,\emph{t}$_{2}$,\emph{t}$_{3}$,\emph{t}$_{4}$,\emph{t}$_{5}$,\emph{t}$_{6}$,\emph{t}$_{7}$,\emph{t}$_{8}$,\emph{t}$_{9}$ can be fired. If \emph{t}$_{1}$was fired, then the collected parameter place \emph{P}$_{B1}$ will get a token which means the parameter of B is obtained, meanwhile, \emph{P}$_{B}$ also get a token and transition \emph{t}$_{1}$ will be fired again. \emph{t}$_{1}$ was fired several times, which is a loop structure represents the process of collecting data several times. When the amount of stimulation of  \emph{t}$_{1}$ is equal to \emph{W}$_{v1}$, the rewritable edge \emph{f}=(\emph{t}$_{1}$,\emph{P}$_{B}$) will disappear. Naturally, when transition \emph{t}$_{2}$ was fired, the acidity collecting parameter places of \emph{P}$_{A1}$ and \emph{P}$_{A2}$ will get a token respectively at the same time. In addition, the acidity parameter place \emph{P}$_{A}$ also get a token and transition \emph{t}$_{2}$ will be fired again. When the times that \emph{t}$_{2}$ was fired is equal to \emph{W}$_{v2}$, the rewritable edge  \emph{f}=(\emph{t}$_{2}$,\emph{P}$_{A}$) will disappear. If the number of tokens in \emph{P}$_{B1}$ and \emph{P}$_{A1}$ is equal to weight function \emph{W}$_{1}$ and \emph{W}$_{2}$ respectively, then the study transition \emph{t}$_{10}$ will be fired. Next, the place \emph{P}$_{BA}$ of relation model for parameter B and A will get a token. When the collect transition \emph{t}$_{3}$  was fired, places \emph{P}$_{M1}$, \emph{P}$_{M21}$ will get a token respectively. Once \emph{t}$_{3}$ is fired and the amount of stimulation is equal to the rewritable constraint \emph{W}$_{v3}$, the rewritable edge \emph{f}=(\emph{t}$_{3}$, \emph{P}$_{M}$) will disappear. If the collection times is equal to the limitation of the weight function \emph{W}$_{v4}$, the collecting parameter place \emph{P}$_{M1}$ will get the amount of tokens of \emph{W}$_{v4}$, at the same time, if the place \emph{P}$_{A2}$ also get the amount of tokens of  \emph{W}$_{v3}$, then the study transition of \emph{t}$_{11}$ can be fired and the relation model place \emph{P}$_{AM}$ of parameter A and M will obtain a token. Similarly, the study transition of \emph{t}$_{12}$, \emph{t}$_{13}$ and \emph{t}$_{14}$ are fired in the same way. Finally, we acquire the prediction model of liquor quality and yield. At this time, the workflow model of the system framewok is completed. In addition, we add an adjust transition \emph{t}$_{16}$ and the edges of (\emph{P}$_{1}$,\emph{t}$_{16}$) and (\emph{t}$_{16}$,\emph{P}$_{start}$) form an extended network which can be used to readjust the precondition parameters \textbf{B}, \textbf{A}, \textbf{M}, etc. 
\begin{figure}
	\centering
	\includegraphics[width=0.5\textwidth, height=8cm]{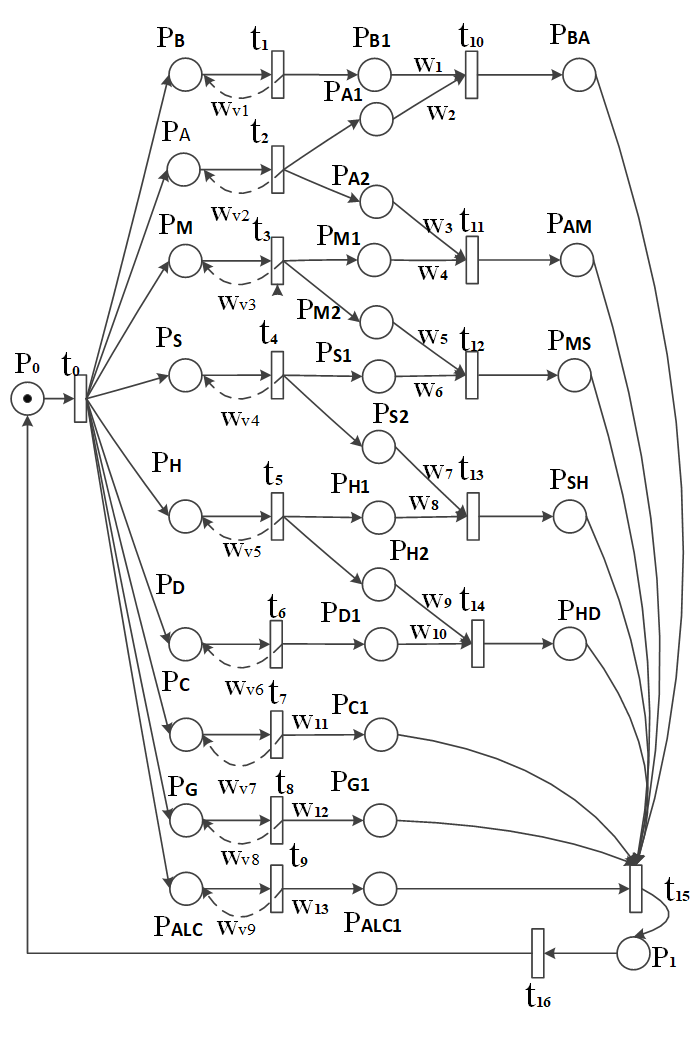}
	\caption{A rewritable petri net model of the system framework.}
	\label{fig:example}
\end{figure}

\newcommand{\tabincell}[2]{\begin{tabular}{@{}#1@{}}#2\end{tabular}}  
\begin{table}[htbp]
	\centering
	\caption{\label{tab:test}The meaning of each place in Fig. 3}
	\begin{tabular}{p{2cm}|p{6cm} }
		\hline
		\tabincell{c}{Places} & Meaning \\
		\hline
		\emph{P}$_{start}$ &  Start\\
		\hline
		\emph{P}$_{B}$, \emph{P}$_{A}$,  \emph{P}$_{M}$, \emph{P}$_{S}$ &  The parameter of \textbf{B}ran shell quality, \textbf{A}cidity, \textbf{M}aternal draff quality and \textbf{S}tarch quality respectively.\\
		\hline
		\emph{P}$_{H}$, \emph{P}$_{D}$,  \emph{P}$_{C}$, \emph{P}$_{G}$, \emph{P}$_{ALC}$  & The parameter of \textbf{H}umidity, \textbf{D}aqu quality, the \textbf{C}ellar-temperature, \textbf{G}round temperature, \textbf{ALC}lcohol concentration respectively.\\
		\hline
		\emph{P}$_{B1}$, \emph{P}$_{C1}$,  \emph{P}$_{G1}$, \emph{P}$_{ALC1}$  &The collected parameter
		of \textbf{B}ran shell quality, the original \textbf{C}ellar-temperature, ground temperature, alcohol concentration respectively.\\
		\hline
		\emph{P}$_{A1}$, \emph{P}$_{A2}$  &The collected parameter of \textbf{A}cidity\\
		\hline
		\emph{P}$_{M1}$, \emph{P}$_{M2}$  & The collected parameter of \textbf{M}aternal draff\\
		\hline
		\emph{P}$_{S1}$, \emph{P}$_{S2}$  & The collected parameter of \textbf{S}tarch quality\\
		\hline
		\emph{P}$_{D1}$, \emph{P}$_{D2}$  & The collected parameter of \textbf{H}umidity\\
		\hline
		\emph{P}$_{BA}$ & The relational model of \textbf{B}ran shell quality and \textbf{A}cidity\\
		\hline
		\emph{P}$_{AM}$ & The relational model of \textbf{A}cidity and \textbf{M}aternal draff quality\\
		\hline
		\emph{P}$_{MS}$ & The relational model of \textbf{M}aternal draff quality and \textbf{S}tarch quality\\
		\hline
		\emph{P}$_{SH}$ & The relational model of \textbf{S}tarch quality and \textbf{H}umidity\\
		\hline
		\emph{P}$_{HD}$ & The relational model of \textbf{H}umidity and \textbf{D}aqu quality\\
		\hline
		\emph{P}$_{1}$ & The relationship model of quality and yield of \textbf{ALC}lcohol\\
		\hline
	\end{tabular}
\end{table}

\begin{table}[htbp]
	\centering
	\caption{\label{tab:test}The meaning of each transition and weight function in Fig. 3}
	\begin{tabular}{p{1.5cm}|p{1cm}|p{1.5cm}|p{3cm} }
		\hline
		\tabincell{c}{T} & Meaning  & W & meaning\\
		\hline
		\emph{t}$_{0}$ &  Into the cellar & \emph{W}$_{v1}$$ - $ \emph{W}$_{v9}$ &  The weight function of rewriteable edge\\
		\hline
		\emph{t}$_{1}$$ - $ \emph{t}$_{9}$ &  collect & \emph{W}$_{1}$$ - $ \emph{W}$_{10}$ &  The weight function \\
		\hline
		\emph{t}$_{10}$$ - $ \emph{t}$_{15}$ &  study &   & \\
		\hline
		\emph{t}$_{16}$ &  adjust &   & \\
		\hline
	\end{tabular}
\end{table}

\subsection{The soundness of the system model}

The soundness of the system model guarantee the correctness of the logics of the system framework. If a petri net is soundness, it will satisfy the following definition.

\textbf{Definition 4} (boundedness [16]) A 7- tuple \emph{EN}=(\emph{P}, \emph{T}, \emph{F}, \emph{K}, \emph{W}, \emph{M}, \emph{W}$_{v}$) is an edge rewritable petri net, where \emph{M}$_{0}$ is the initial marking. If $\exists$ \emph{B} $\in$  $\mathbb{N}$$^{*}$ and $\forall$ \emph{M} $\in$ \emph{R}(\emph{M}$_{0}$): \emph{M}$_{p}$ $\leq$ \emph{B}, the place p is bounded. The minimum \emph{B} is the bound of place \emph{p}, which is denoted as \emph{B}(\emph{p}). If each \emph{p} $\in$\emph{P} in the net is bounded, \emph{EN} is a bounded petri net.

\emph{B}(\emph{p})=min\{\emph{B} $\mid $ $\forall$ \emph{M} $\in$ \emph{R}(\emph{M}$_{0}$): \emph{M}(\emph{p}) $\leq$ \emph{B}\}

\textbf{Definition 5} (liveness [16]) The edge rewritable petri net \emph{EN}=(\emph{P}, \emph{T}, \emph{F}, \emph{K}, \emph{W}, \emph{M}, \emph{W}$_{v}$), if \emph{t} $\in$
\emph{T} and  $\forall$ \emph{M} $\in$ \emph{R}(\emph{M}$_{0}$), $\exists$ \emph{M}$^{\prime}$[\emph{t}$\rangle$, the transition t is liveness. If $\forall$ \emph{t} $\in$ \emph{T} is liveness, the edge rewritable petri net \emph{EN} is liveness.

\textbf{Definition 6} (soundness [16]) \emph{EN}=(\emph{P}, \emph{T}, \emph{F}, \emph{K}, \emph{W}, \emph{M}, \emph{W}$_{v}$) is soundness if the following conditions hold:

(1) For each marking \emph{M} which is reached by the start marking \emph{M}$_{0}$ (the initial place \emph{p}$_{start}$ contains a token), there have transition sequence $\sigma$$_{1}$  and $\sigma$$_{2}$ that the end marking \emph{M}$_{end}$ reached from marking \emph{M}, that is $\forall$\emph{M}: \emph{M}$_{0}$[$\sigma$$_{1}$ $\rangle$ \emph{M} $\rightarrow$ \emph{M}[$\sigma$$_{1}$ $\rangle$ \emph{M}$_{end}$.

(2)  The end marking \emph{M}$_{end}$(the end place \emph{p}$_{end}$ contains a token) is reachable from \emph{M} and \emph{M}$_{end}$ is the only one end marking, that is $\forall$\emph{M}: (\emph{M}$_{0}$[$\sigma$$\rangle$ \emph{M} $\cap$ \emph{M} $\geq$\emph{M}$_{end}$)$\rightarrow$(\emph{M}=\emph{M}$_{end}$).

(3) There is no dead marking (Marking M can not be changed) in \emph{EN}, that is $\forall$\emph{t}$\in$ \emph{T},  $\exists$\emph{M} and \emph{M}$^{\prime}$ have \emph{M}$_{0}$[$\sigma$ $\rangle$ \emph{M} [\emph{t}$\rangle$ \emph{M}$^{\prime}$.

\textbf{Theorem 1} The edge rewritable petri net \emph{EN} is soundness if the extended workflow [17] of \emph{EN} is liveness and boundedness[16].

Reachability graph is a main tool to analysis the petri nets[16], which can be used to analyzed the boundness and liveness of petri net. Each point is a making in the reachability graph. We use the analysis software of petri net to build the reachability graph for the system rewritable petri net model. The result is that the reachability graph includes 1866 reachability status and 8750 arcs. However, this reachability graph is too huge to show visually,and the result of analysis software is shown in Fig. 4. Naturally, the reachability graph need to be compressed, which satisfy the principle that the paths are not lost. 
\begin{figure}
	\centering
	\includegraphics[width=0.4\textwidth, height=4cm]{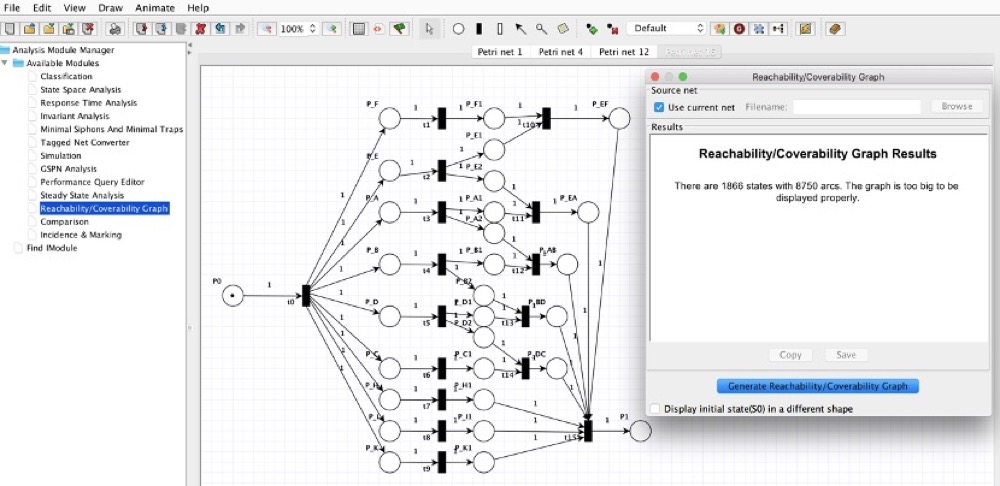}
	\caption{The analysis of reachability graph of the rewritable petri net model of system framework}
	\label{fig:example}
\end{figure}

\textbf{Definition 7} (Homogenous sequence) $\sigma$$_{1}$  and $\sigma$$_{2}$ are the transition sequences, if $\emph{M} \xrightarrow{\sigma_{1}}$ \emph{M}$^{\prime}$, $\emph{M} \xrightarrow{\sigma_{2}}$ \emph{M}$^{\prime}$, $\forall$ \emph{t}$\in$ $\sigma$$_{1}$ $\land$ $\forall$ \emph{t}$\in$$\sigma$$_{2}$ $\rightarrow$ \#(\emph{t}/$\sigma$$_{1}$) = \#(\emph{t}/$\sigma$$_{2}$), then $\sigma$$_{1}$ and $\sigma$$_{2}$ are the homogenous sequence.  Using ($\sigma$$_{1}$) or ($\sigma$$_{2}$) to represent homogenous sequence, and \#(\emph{t}/$\sigma$$_{1}$), \#(\emph{t}/$\sigma$$_{2}$) denote the number of occurrences of transition t in $\sigma$$_{1}$ and $\sigma$$_{2}$.

\textbf{Property 1: } According to definition 3, Two homogeneous sequence start from the same marking \emph{M} and reaches to another equal marking \emph{M}$^{\prime}$ through different transition sequences. So, the paths between \emph{M} and \emph{M}$^{\prime}$ can be compressed, and the reachability of the original reachability graph remains unchanged after compression.

We compress the reachability graph of the system rewritable petri nets model according to the principle of homogeneous sequence compression. Because of the transition \emph{t}$_{1}$, \emph{t}$_{2}$, \emph{t}$_{3}$, \emph{t}$_{4}$, \emph{t}$_{5}$, \emph{t}$_{6}$, \emph{t}$_{7}$, \emph{t}$_{8}$, \emph{t}$_{9}$ will be fired for many times according to the collection times in the system rewritable petri nets model. Using {\emph{t}$_{i}$}$^{*}$ denotes the transition sequence \emph{t}$_{i}$\emph{t}$_{i}$...\emph{t}$_{i}$ that \emph{t}$_{i}$ has been fired many times continuously. Meanwhile, transition sequence {\emph{t}$_{7}$}$^{*}$, {\emph{t}$_{8}$}$^{*}$, {\emph{t}$_{9}$}$^{*}$ are compressed as ($\sigma$), and the reachability graph of the compressed system model is shown in Fig. 5. The transition sequence between the markings in Fig. 5 is shown in Table III.

\begin{figure}
	\centering
	\includegraphics[width=0.5\textwidth]{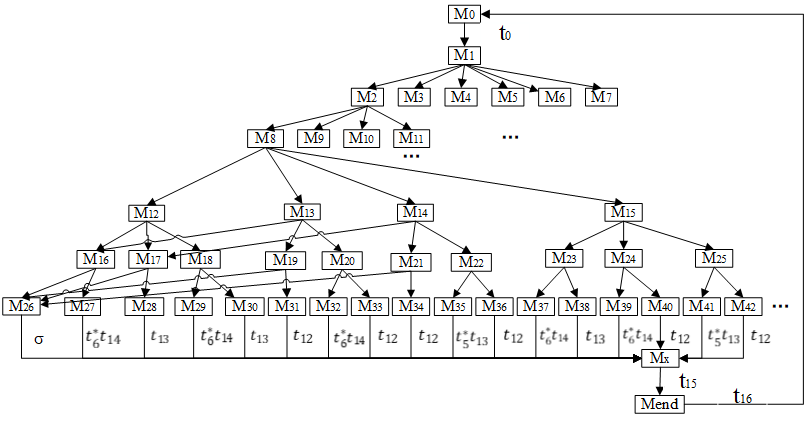}
	\caption{The compressed reachability graph of the rewritable petri net model of system framework}
	\label{fig:example}
\end{figure}

\renewcommand{\arraystretch}{1.5} 
\begin{table}
	\centering
	\fontsize{6.5}{8}\selectfont
	\begin{threeparttable}
		\caption{Transition sequences between the markings}
		\label{tab:performance_comparison}
		\begin{tabular}{ccccc}
			\midrule
			$M_1 \rightarrow {M_2}$ & $ ({t_1}^*{t_2}^*)t_{10} $    & $M_{12} \rightarrow {M_{16}}$ & $ {t_5}^*t_{13} $          \cr
			$M_1 \rightarrow {M_3}$ & $ ({t_2}^*{t_3}^*)t_{11} $    & $M_{12} \rightarrow {M_{17}}$ & $ ({t_5}^*{t_6}^*)t_{14} $           \cr
			$M_1 \rightarrow {M_4}$ & $ ({t_3}^*{t_4}^*)t_{12} $    & $M_{12} \rightarrow {M_{18}}$ & $ (\sigma)$                   \cr
			$M_1 \rightarrow {M_5}$ & $ ({t_4}^*{t_5}^*)t_{13} $    & $M_{13} \rightarrow {M_{16}}$ & $ t_{12} $                         \cr
			$M_1 \rightarrow {M_6}$ & $ ({t_5}^*{t_6}^*)t_{14} $    & $M_{13} \rightarrow {M_{19}}$ & $ {t_6}^*t_{14} $                     \cr
			$M_1 \rightarrow {M_7}$ & $ (\sigma)$                   & $M_{13} \rightarrow {M_{20}}$ & $ (\sigma)$                                \cr
			$M_2 \rightarrow {M_8}$ & $ {t_3}^*t_{11} $             & $M_{14} \rightarrow {M_{17}}$ & $ t_{12} $                                \cr
			$M_2 \rightarrow {M_9}$ & $ ({t_3}^*{t_4}^*)t_{12} $    & $M_{14} \rightarrow {M_{21}}$ & $ {t_5}^*t_{13} $                    \cr
			$M_2 \rightarrow {M_{10}}$ & $ ({t_4}^*{t_5}^*)t_{13} $ & $M_{14} \rightarrow {M_{22}}$ & $ (\sigma)$                                 \cr
			$M_2 \rightarrow {M_{11}}$ & $ ({t_5}^*{t_6}^*)t_{14} $ & $M_{15} \rightarrow {M_{23}}$ & $ {t_4}^*t_{12} $                         \cr
			$M_8 \rightarrow {M_{12}}$ & $ {t_4}^*t_{12} $          & $M_{15} \rightarrow {M_{24}}$ & $ ({t_4}^*{t_5}^*)t_{13} $                 \cr
			$M_8 \rightarrow {M_{13}}$ & $ ({t_4}^*{t_5}^*)t_{13} $ & $M_{15} \rightarrow {M_{25}}$ & $ ({t_5}^*{t_6}^*)t_{14} $          \cr
			$M_8 \rightarrow {M_{14}}$ & $ ({t_5}^*{t_6}^*)t_{14} $ & $M_{16} \rightarrow {M_{26}}$ & $ {t_6}^*t_{14} $                   \cr
			$M_8 \rightarrow {M_{15}}$ & $ (\sigma)$                & $M_{16} \rightarrow {M_{27}}$ & $ (\sigma)$                 \cr
			
			$M_{17} \rightarrow {M_{26}}$ & $ t_{13} $                 & $M_{24} \rightarrow {M_{39}}$ & $ t_{12} $   \cr
			$M_{17} \rightarrow {M_{28}}$ & $ (\sigma)$                & $M_{24} \rightarrow {M_{40}}$ & $ {t_6}^*t_{14} $ \cr
			$M_{18} \rightarrow {M_{29}}$ & $ {t_5}^*t_{13} $          & $M_{25} \rightarrow {M_{41}}$ & $ t_{12} $ \cr
			$M_{18} \rightarrow {M_{30}}$ & $ {t_6}^*t_{14} $          & $M_{25} \rightarrow {M_{42}}$ & $ {t_5}^*t_{13} $ \cr
			$M_{19} \rightarrow {M_{26}}$ & $ t_{12} $ \cr
			$M_{19} \rightarrow {M_{31}}$ & $ (\sigma)$ \cr
			$M_{20} \rightarrow {M_{32}}$ & $ t_{12} $  \cr
			$M_{20} \rightarrow {M_{33}}$ & $ {t_6}^*t_{14} $ \cr
			$M_{21} \rightarrow {M_{26}}$ & $ t_{12} $ \cr
			$M_{21} \rightarrow {M_{34}}$ & $ (\sigma)$ \cr
			$M_{22} \rightarrow {M_{35}}$ & $ t_{12} $ \cr
			$M_{22} \rightarrow {M_{36}}$ & $ {t_5}^*t_{13} $ \cr
			$M_{23} \rightarrow {M_{37}}$ & $ {t_5}^*t_{13} $ \cr
			$M_{23} \rightarrow {M_{38}}$ & $ ({t_5}^*{t_6}^*)t_{14} $ \cr
			\bottomrule
		\end{tabular}
	\end{threeparttable}
\end{table}

The original workflow will becomes an extended workflow when adding a new transition \emph{t}$_{16}$ in Fig. 3. If \emph{t}$_{16}$ is fired, the end marking will be transformed to the initial marking \emph{M}$_{0}$. Since $\forall$ \emph{t}$\in$ \emph{T} and $\forall$ \emph{M}$\in$ \emph{R}(\emph{M}$_{0}$) are all $\exists$ \emph{M}$^{\prime}$ $\in$ \emph{R}(\emph{M}), \emph{M}$^{\prime}$[\emph{t}$\rangle$, the extended workflow with \emph{t}$_{16}$ is liveness. The rewritable edge such as (\emph{t}$_{1}$, \emph{P}$_{F}$), (\emph{t}$_{2}$, \emph{P}$_{E}$), (\emph{t}$_{3}$,\emph{P}$_{B}$) etc. will disappear, which depend on the number of collection times. The rewritable edge guarantees the boundedness of the places in the system framework model of rewritable petri net. 

In summary, the extended work flow is liveness and boundedness through the analysis. Therefore, the system framework model are soundness. 

\section{Our proposed IUP Scheme}
In order to realize the utility prediction, we use deep learning to obtain the relationship between the parameters of the SSF process. To achieve sufficient data, this Section introduces the data generator model of MSE-GAN and the utility prediction model of SSF with fully connected networks.
\subsection{The model of GAN} 
As introduced in the related work, we use GAN to generate more effective data to solve the few-shot problem. At the beginning of the training, keep the generator unchanged and then train the discriminator. Then, the generator and the discriminator play games. When the performance of generator is no longer improved, next, keep the discriminator unchanged and train the generator. The data generated by the generator, then the discriminator distinguish the real data and the data generated with the generator. Through the constant game between the discriminator and the generator. Finally, the discriminator cannot distinguish the real data and the generated data. The original GAN generate data algorithm is shown in [22].
\begin{figure}
	\centering
	\includegraphics[width=0.48\textwidth, height=5cm]{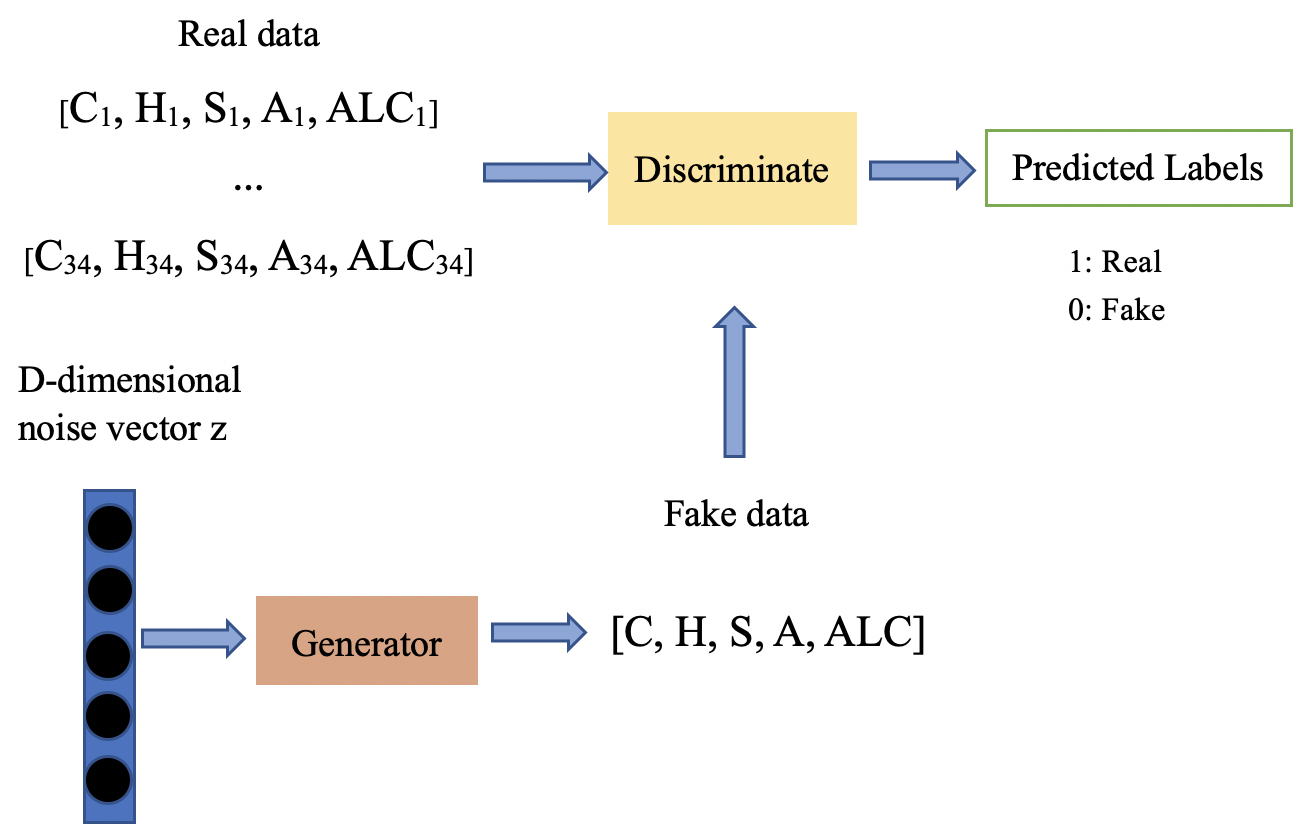}
	\caption{The generation model of one-dimensional few-shot data based on GAN}
	\label{fig:example}
\end{figure}

\subsection{The data generation model of few-shot one-dimensional based on MSE-GAN}

The generator of MSE-GAN is a  FCNN with four-layer, which input layer is five neurons. The output layer of generator has  five neurons which is corresponding to the five categories of the generated one-dimensional data. The discriminator with three layers of fully connected network, which input is the one-dimensional data of five categories and the output is the data of 0 to 1 which represent the evaluation of whether the input data is real data. The network structure of generator and discriminator are shown in Fig. 7.

\begin{figure}[htbp]
	\centering\includegraphics[width=\linewidth]{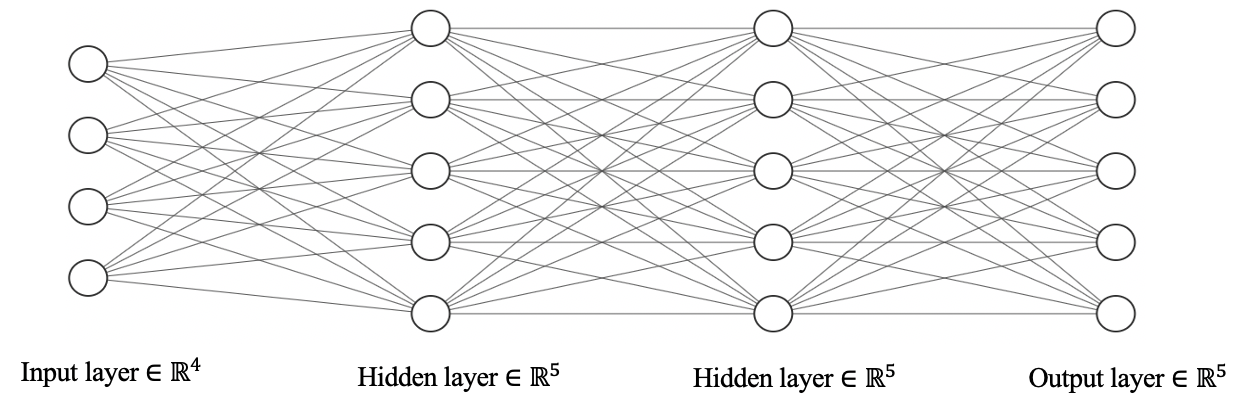}
	\centerline{(a) generator}
	\\
	\centerline{\includegraphics[width=\linewidth]{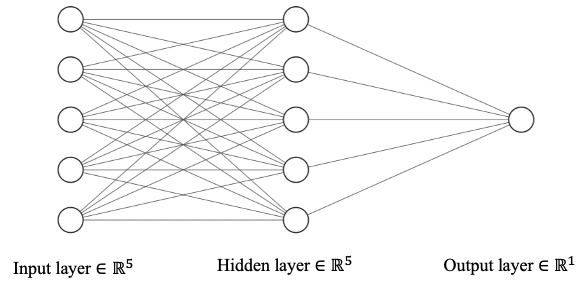}}
	\centerline{(b) discriminator}
	\caption{The network structure of generator and discriminator}
	\label{fig:res}
\end{figure}

To avoid the large difference between the real data and generated data which is generated by the original GAN. We invent the algorithm of mean square error(MSE) with GAN to generate more effective one-dimensional data. The process is as follows, first, the real data is used as the input training data set, and then repeat training the 5 epochs by the model in Fig. 6, this step will generate amount data. Finally, making MSE between the generated data and the original data, as well as the generated data with a MSE greater than the threshold value are filtered out. The remaining data are retained as the appropriate generated data. The algorithm is as follows:

\IncMargin{1em}
\begin{algorithm}
	
	\SetAlgoNoLine
	\SetKwInOut{Input}{\textbf{Input}}\SetKwInOut{Output}{\textbf{Output}}
	\Input{
		Generated(original) data set; Real data set;\\}
	\Output{
		Generated(final) data set;\\}
	\BlankLine
	\For{data(g) in Generated(original) data set}{
		\For{data(r) in real data set}{
			\If{The MSE of data(g)and data(r) $\leq$ threshold value}{ 
				Put the data(g) into Generated(final) data set.\\
				Break.
				
			}
		}
	}
	
	\caption{The algorithm of MSE-GAN \label{al2}}
\end{algorithm}

\subsection{The utility prediction model for Chinese liquor SSF with fully connected neural network}

This paper uses the FCNN to model the utility prediction of Chinese liquor SSF, the model is shown in Fig. 8. The main component of maternal draff is starch. Daqu is mainly composed of microorganisms. Bran shell provides more aerobic breathing space for microbial metabolism. Temperature is a necessary condition for SSF. As a result, the main parameters of SSF process are humidity, starch, acidity and temperature. These four parameters affect the quality and yield of alcohol. Therefore, firstly, we use temperature, humidity, starch and acidity as input vector, which dimension is 1$\times$4. Secondly, the input vector goes through the 4$\times$64 dimensions, 64$\times$128 dimensions, 128$\times$256 dimensions and 256$\times$128 dimensions orderly to explore the underlaying connections between the input data. Finally, we can use a single neural getting one final value. The final production of alcohol value is activated by Tanh(x) function which mitigate the problem of gradient disappearance. A batch normalization operation is added to ensure the stable of each output layer and reduce the phenomenon of gradient disappearance.

\begin{figure}[htb]
	\centering
	\includegraphics[width=0.5\textwidth]{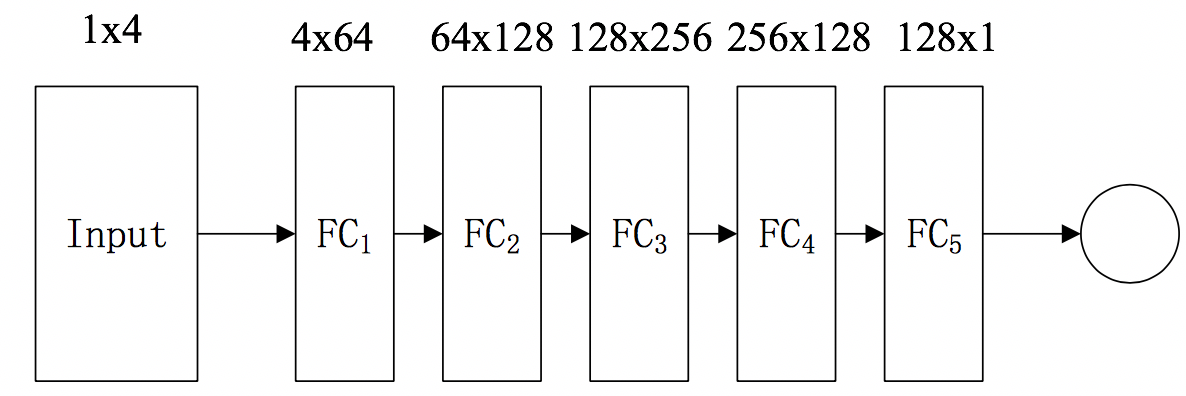}
	\caption{A regression model with FCNN}
	\label{fig:example}
\end{figure}
In this paper, we design the model of utility prediction for SSF, which includes learning algorithm and prediction algorithm. 

Step 1: learning algorithm include 5 main input parameters Alcohol (\emph{ALC}), cellar temperature (\emph{C}), Humidity (\emph{H}), Starch content (\emph{S}), and Acidity (\emph{A}). In the learning algorithm, we use the mean-square error as the loss function which is used to control the end of the loop and get the most accurate learning weight W. According to the gradient descent function and the learning rate, the learning weight function W is obtained.

Step 2: according to the learning algorithm, The relationship model \emph{F}$_{w}$ is acquired. The model of \emph{F}$_{w}$ is a trained FCNN, which has been learned the relationships between parameters. Input the parameters of cellar temperature (\emph{C}), Humidity (\emph{H}), Starch content (\emph{S}), and Acidity (\emph{A}) into the model will get the target value Alcohol (\emph{ALC}). These two algorithms are as follows.

	\renewcommand{\algorithmicrequire}{ \textbf{Input:}} 
	\renewcommand{\algorithmicensure}{ \textbf{Output:}} 
	\begin{algorithm}[htb] 
		\caption{ The learning algorithm:.} 
		\label{alg:Framwork} 
		\begin{algorithmic}[1] 
			\REQUIRE ~~\\ 
			\emph{C},\emph{H},\emph{S},\emph{A}, \emph{ALC};
			\ENSURE ~~\\ 
			W;
			\WHILE{ Loss=(F(\emph{C},\emph{H},\emph{S},\emph{A})-\emph{ALC})$^2$ $>$ threshold}
			\STATE $\Delta \leftarrow$ -\emph{g}((\emph{F}(\emph{C},\emph{H},\emph{S},\emph{A})-\emph{ALC})$^2$)\qquad     (Gradient descent) 
			\STATE \emph{W} = \emph{W}+ $\alpha\Delta$ \qquad             ( $\alpha$ is the  learning rate )             
			\ENDWHILE
			\RETURN \emph{W}; 
		\end{algorithmic}
	\end{algorithm}
	
	\renewcommand{\algorithmicrequire}{ \textbf{Input:}} 
	\renewcommand{\algorithmicensure}{ \textbf{Output:}} 
	\begin{algorithm}[htb] 
		\caption{ The prediction algorithm:.} 
		\label{alg:Framwork} 
		\begin{algorithmic}[1] 
			\REQUIRE ~~\\ 
			\emph{C},\emph{H},\emph{S},\emph{A};
			\ENSURE ~~\\ 
			\emph{ALC};
			\STATE \emph{ALC} $\leftarrow$ \emph{F}$_{w}$(\emph{C},\emph{H},\emph{S},\emph{A})\\ \qquad   (\emph{F}$_{w}$ is the relationship model of the  process parameter) 
			\RETURN \emph{ALC}; 
		\end{algorithmic}
	\end{algorithm}
	
	\subsection{Data preprocessing}
	The data set is divided into training data set and test data set according to the ratio of 4:3. Partial SSF data are shown in Table IV. We can see that the range of values for each dimension attribute varies greatly. In order to accelerate the speed of network convergence, we adopt a min-max method to scale all the original data into [0,1], the method is described as follows:
	\begin{center}
		\emph{X}=$\frac{\emph{X}-\emph{Min}}{\emph{Max}-\emph{Min}}$.
	\end{center}
	
	\begin{table}[htbp]
		\centering
		\caption{\label{tab:test}The raw parameters of SSF}
		\begin{tabular}{cccc}
			\hline
			\tabincell{c}{ Cellar Temperature} & Humidity\%  & Starch content\% & Acidity\%\\
			\hline
			40&  45.31 & 35.17 &  1.42\\
			\hline
			40.5&  45.4 & 34.85 &  1.55 \\
			\hline
			43 &  45.4 &  34.25 & 1.69\\
			\hline
			42 &  45.42 & 34.12  &1.69 \\
			\hline
		\end{tabular}
	\end{table}
	
	\section{Performance Analysis}
	In this Section, the accuracy of the prediction is tested and we compare our method with other prediction means. In addition,  we also compare the predicted time with other methods
	\subsection{Experimental data generation}
	There are several parameters take effects on the process of liquor solid fermentation: temperature  (\emph{C}), humidity (\emph{H}), starch (\emph{S}) and acidity (\emph{A}). We take them to be the main measurable attributes about the production of alcohol (\emph{ALC}). Because other attributes of material carry out very limited influence to the alcohol, we will not take them into consideration. All of our experimental data come from the three wine cellars with similar environment A1, A2 and A3. We record J D B and E every two days and get 11, 11 and 12 groups of data from these wine cellars. Each pace of data is 1 * 5 vectors.
	
	First, we add some random factors to achieve the enhanced data, and put them into the original GAN to generate experimental data. However, the generated data differs greatly from the real data by original GAN. After that, we use the MSE-GAN to generate data under the same condition, which includes the same neural network and iterations. The data generated by the threshold value of 0.15 will be closer to the original data through several experiments. The results are shown in Fig. 9.
	
	\begin{figure}
		\begin{minipage}{0.48\linewidth}
			\centerline{\includegraphics[width=4.0cm]{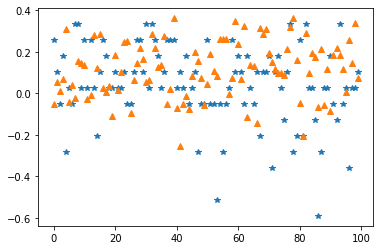}}
			\centerline{(a) C}
		\end{minipage}
		\hfill
		\begin{minipage}{.48\linewidth}
			\centerline{\includegraphics[width=4.0cm]{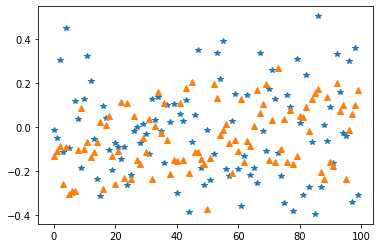}}
			\centerline{(b) H}
		\end{minipage}
		\vfill
		\begin{minipage}{0.48\linewidth}
			\centerline{\includegraphics[width=4.0cm]{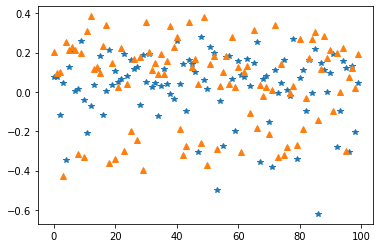}}
			\centerline{(c) S}
		\end{minipage}
		\hfill
		\begin{minipage}{0.48\linewidth}
			\centerline{\includegraphics[width=4.0cm]{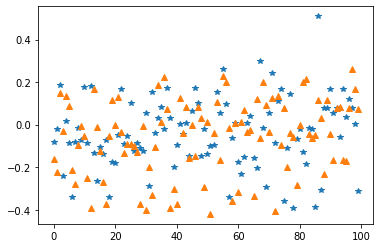}}
			\centerline{(d) A}
		\end{minipage}
		\vfill
		\begin{minipage}{0.48\linewidth}
			\centerline{\includegraphics[width=4.0cm]{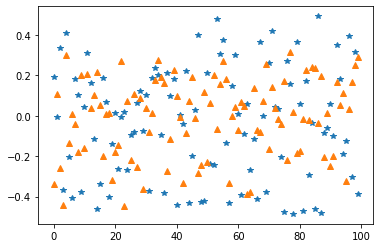}}
			\centerline{(d) ALC}
		\end{minipage}
		
		\caption{The comparison of real data and data generated by MSE-GAN(real data: blue, generated data: orange)}
		\label{fig:res}
	\end{figure}
	
	Fig. 9 presents the comparison of real data and generated data of 5 class which include C, H, S, A and ALC by MSE-GAN. we can see that the data generated by the MSE-GAN present a similar trend as the real data. Consequently, the data generated by MSE-GAN can be used for experimental analysis.   
	
	\subsection{Comparison and analysis of experimental results}
	By the approach of MSE-GAN, we generated 1077 cases of data to train the FCNN to obtain the relationship between parameters. The model's structure is shown in Fig. 10, which is a FCNN model with one input layer, four hidden layers and an output layer. Input layer and each hidden layer includes 4 cells. 
	
	\begin{figure}
		\centering\includegraphics[width=\linewidth]{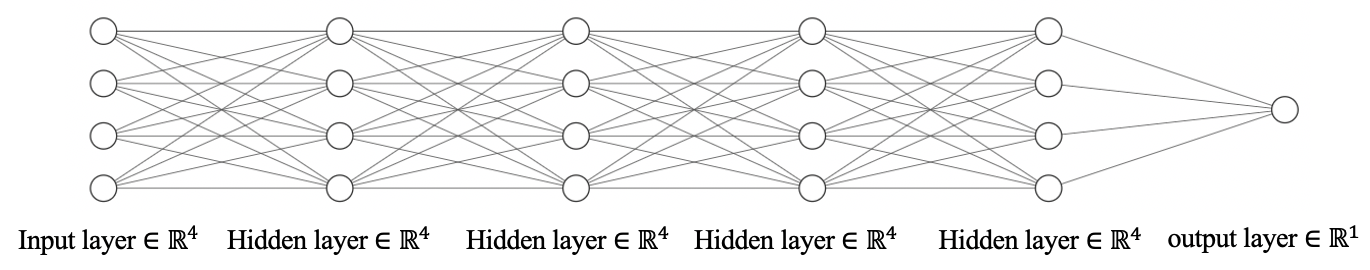}
		\caption{The alcohol prediction model.}\label{fig:r2}
	\end{figure}

	To compare the differences of the original GAN and the MSE-GAN, we use the data which generated by original GAN and MSE-GAN to train the alcohol prediction model of FCNN respectively, and use the 34 groups of real data to validate it. The results are shown in Fig. 11.  The abscissa represents the sample number and the ordinate represents the alcohol concentration. The green line represents the original real alcohol concentration data, and the red line represents the alcohol concentration predicted by the trained FCNN after the generation of the original GAN model and MSE-GAN model respectively.
	
	\begin{figure}
		\begin{minipage}{0.48\linewidth}
			\centerline{\includegraphics[width=4.0cm]{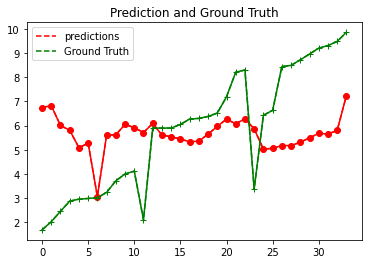}}
			\centerline{a. Original GAN }
		\end{minipage}
		\hfill
		\begin{minipage}{0.48\linewidth}
			\centerline{\includegraphics[width=4.0cm]{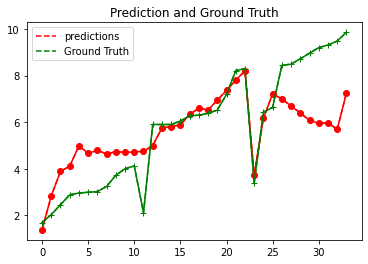}}
			\centerline{b. MSE-GAN}
		\end{minipage}
		\caption{FCNN prediction results of alcohol concentration based on the data generated by the Original-GAN  and MSE-GAN}
		\label{fig:res}
	\end{figure}
	In this paper, we use the MSE to measure the effect of regression. The smaller the mean square error is, the more effective the regression prediction is. The MSE value of the data generated by the original GAN and MSE-GAN are 8.35 and 1.618 respectively, so the effect of data fitting which generated by original GAN is less than that generated by MSE-GAN.
	
	In addition, we compare the method of FCNN with GAN prediction and the Multiple Linear Regression(MLR) by using the data which generated by the MSE-GAN. The prediction results are shown in Fig. 12, and the results of MSE are shown in Table V.

	\begin{figure}
		\begin{minipage}{0.48\linewidth}
			\centerline{\includegraphics[width=4.0cm]{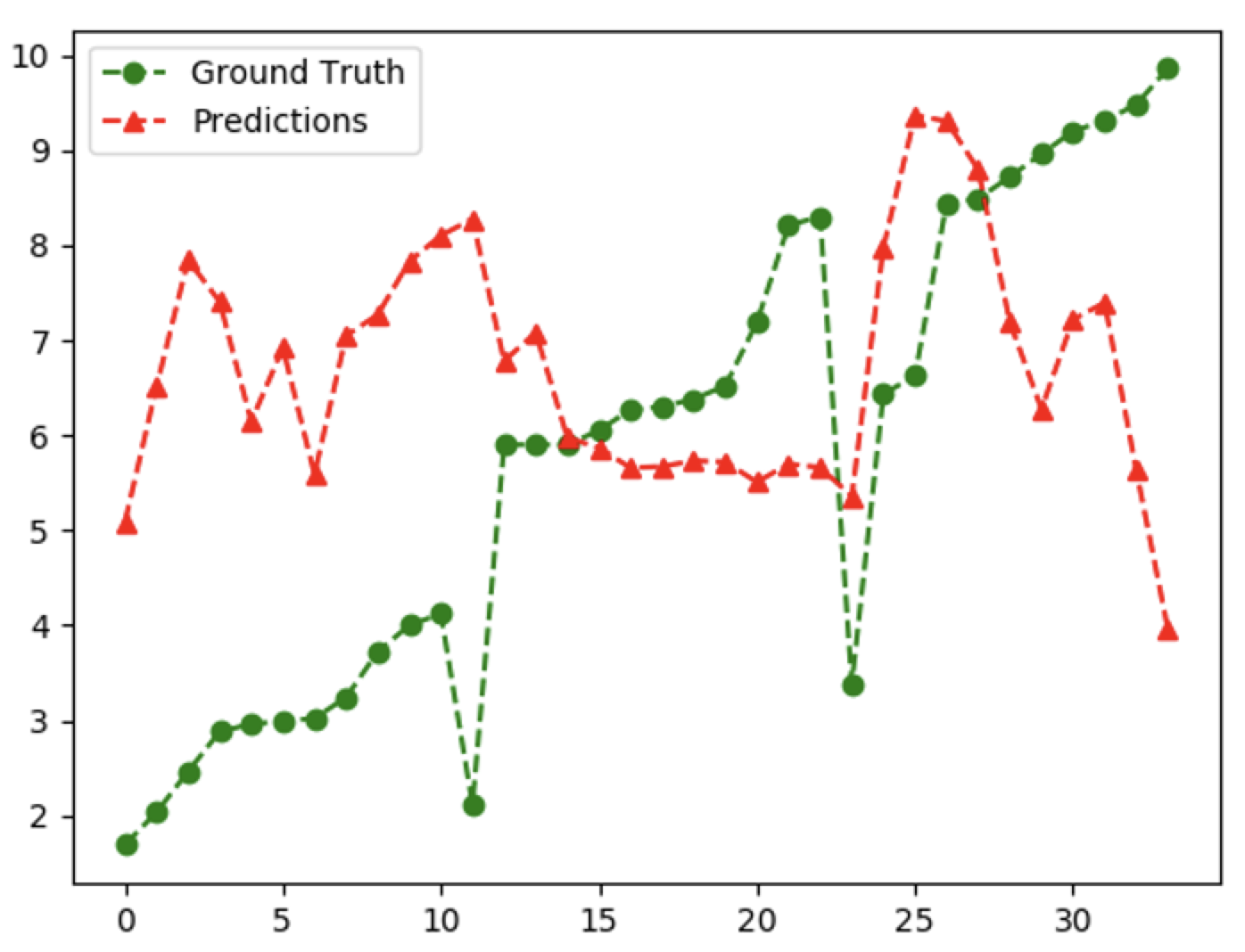}}
			\centerline{a. GAN prediction}
		\end{minipage}
		\hfill
		\begin{minipage}{0.48\linewidth}
			\centerline{\includegraphics[width=4.2cm, height=3cm]{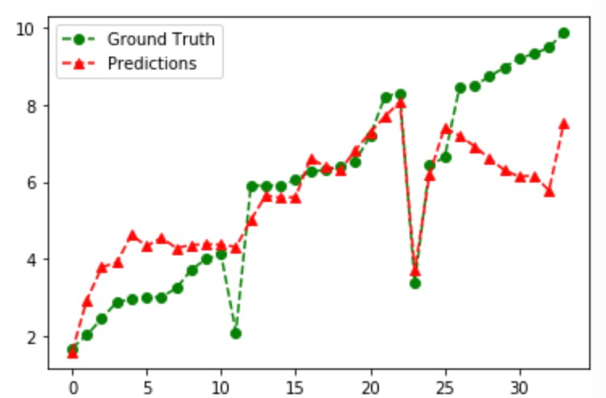}}
			\centerline{b. MLR}
		\end{minipage}
		\caption{The alcohol prediction results based on the prediction and MLR}
		\label{fig:res}
	\end{figure}

	\begin{table}
		\centering
		\caption{\label{tab:test}The mean-square error of three methods}
		\begin{tabular}{cc}
			\hline
			\tabincell{c}{Method} & MSE\\
			\hline
			GAN prediction &  9.128 \\
			\hline
			MLR &  2.167 \\
			\hline
			FCNN & 1.618 \\
			\hline
		\end{tabular}
	\end{table}
	
	Moreover, we compare the prediction time of the three algorithms. We made the predictions for 34, 429, 750 and 1,077 data sets, respectively.  The time increment percentage of the three algorithms is shown in the Fig. 13.
	
	\begin{figure}[H]
		\centering\includegraphics[width=\linewidth]{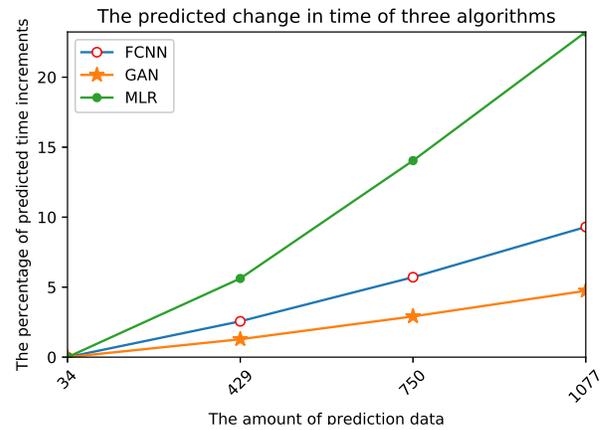}
		\caption{The prediction time performance of three algorithms }\label{fig:r2}
	\end{figure}
	
	It can be seen from the Table V that the MSE of FCNN is minimum, so FCNN is more effective than the other two methods. According to Fig. 13, we can see that the time change percentage of MLR is larger and the time change of GAN prediction is smaller. The difference of time change percentage between FCNN and GAN prediction is less than 5\%.
	
	In summary, the prediction accuracy and time performance of FCNN meet the expectations. This method can predict the quality and yield of SSF accurately in short time.
	
	\subsection{Engineering Application}
	
	SSF has a good application in food industry, enzyme preparation, organic acid flavor and other fields. SSF put the pre-proportion raw materials into the fermentation tank, and the fermentation process can not be controlled or changed any more. The production of liquor mainly relies on SSF. Fig. 14-15 shows the fermentation tank and digital management of a liquor SSF.
	
	\begin{figure}[H]
		\centering\includegraphics[width=0.8\linewidth]{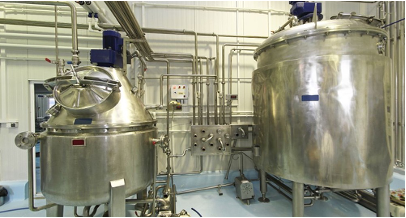}
		\caption{The fermentation tank of liquor SSF}\label{fig:r2}
	\end{figure}
	
	\begin{figure}[H]
		\centering\includegraphics[width=0.9\linewidth]{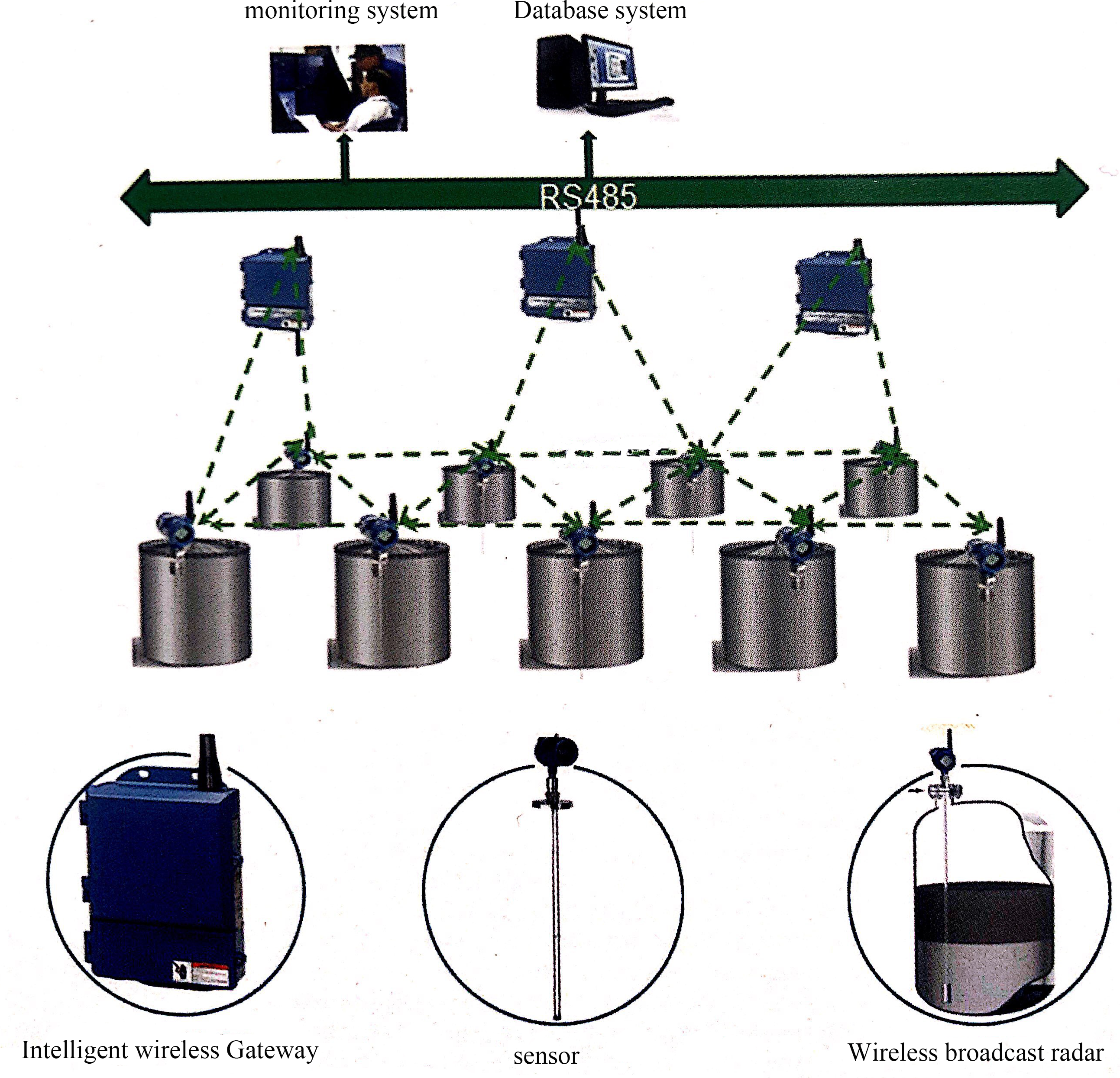}
		\caption{The digital management of liquor SSF}\label{fig:r2}
	\end{figure}
	However, the proportion of raw material depends on the artificial expertise, which would lead to the unstable yield and quality, as well as the waste of grain. To improve the utilization of grain in SSF, we can predict the quality and yield in advance according to the proportion of raw materials before SSF. In Fig. 16, we use deep learning to analyze the data of SSF collected in real time. Specifically, we use the improved GAN algorithm to generate the data when the collecting data is not enough. Then, the generated data and the actual data will be used as the training set by the FCNN for analyzing the relationship of these data. Finally, the yield and quality of SSF can be predicted according to the proportion of raw materials by FCNN.
	
	\begin{figure}
		\centering\includegraphics[width=\linewidth]{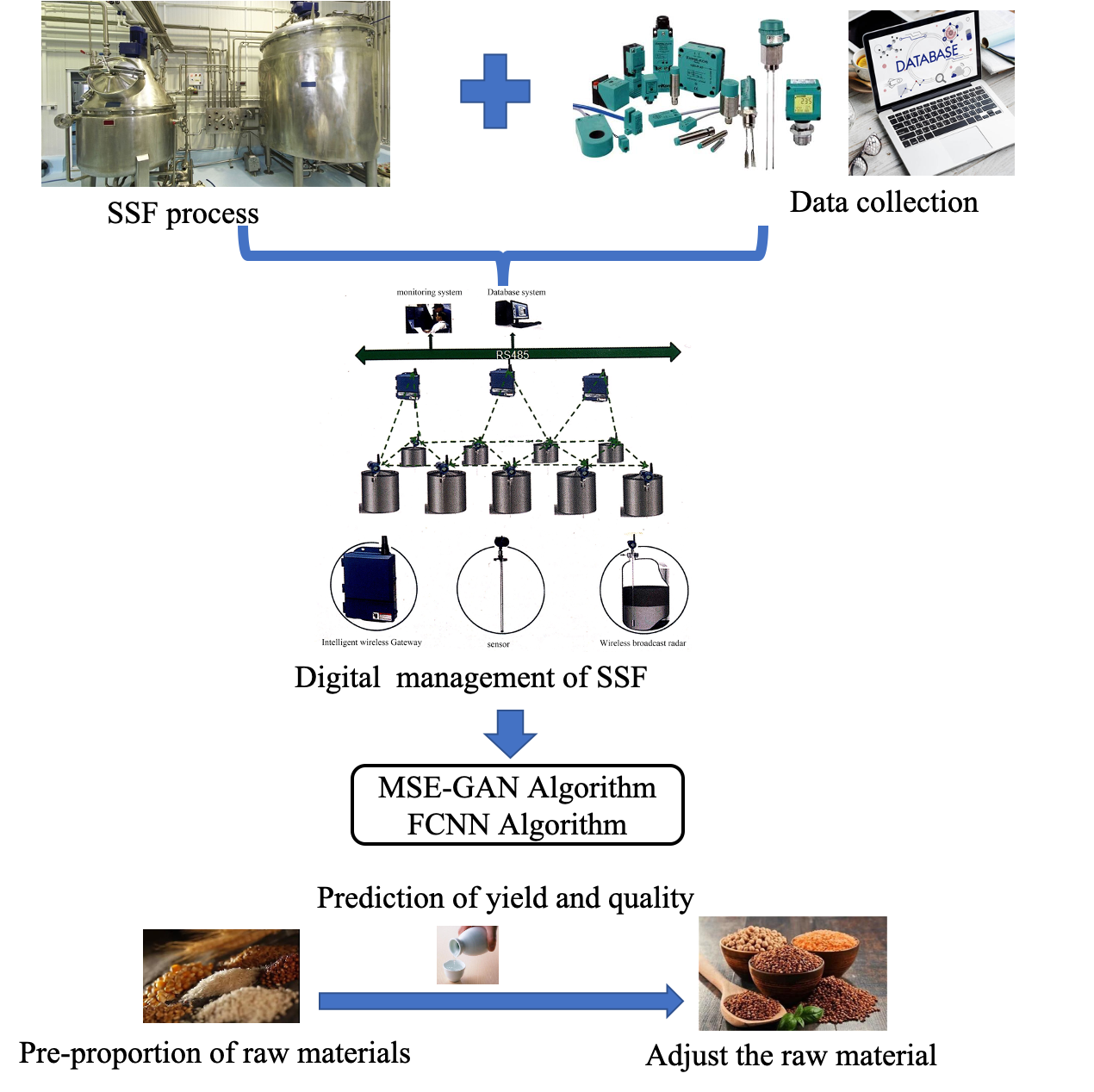}
		\caption{The application of SSF prediction}\label{fig:r2}
	\end{figure}
	
	We will focus on developing more efficient ways to adjust the proportion of raw materials according to the predicted quality and yield. In addition, the influence of microorganisms on SSF was not considered in this paper. In the future, the influence of microorganisms on SSF will be studied to further improve the utilization of grain.

	\section{conclusion and future work}
	We design a system for collecting parameters and predicting the quality and yield of liquor SSF in this paper. The system framework of liquor SSF was modeled by the edge-rewritable petri nets and the soundness of the model was proved. Because the data of SSF  process is not enough, we proposed MSE-GAN to expand the original data as the training data. The relationship between the main parameters of the SSF process is analyzed by using the FCNN. Finally, the alcohol concentration can be effectively predicted according to the parameter of acidity, starch content, humidity and temperature. In this way, this paper  provides data analysis for the ratio of preconditions and improves the yield and quality of liquor.
	
	Since the effects of starch content, humidity, acidity and temperature on alcohol concentration were considered in this paper. However, microorganisms are another factors that affect the quality of SSF. In the future, we will study the quality of SSF by various microorganisms.

\ifCLASSOPTIONcompsoc
  \section*{Acknowledgments}
\else
  \section*{Acknowledgment}
\fi

This work was partially supported by the National Nature Science Foundation of China under Grant Nos.61572523, 61873281, 61672033, 61873280, and 61972416.

\ifCLASSOPTIONcaptionsoff
  \newpage
\fi

\end{document}